\documentclass[letterpaper]{article} % DO NOT CHANGE THIS
\usepackage[preprint]{aaai2027}  % DO NOT CHANGE THIS
\usepackage[hyphens]{url}  % DO NOT CHANGE THIS
\usepackage{graphicx} % DO NOT CHANGE THIS
\usepackage{natbib}  % DO NOT CHANGE THIS AND DO NOT ADD ANY OPTIONS TO IT
\usepackage{caption} % DO NOT CHANGE THIS AND DO NOT ADD ANY OPTIONS TO IT
\usepackage{algorithm}
\usepackage{algorithmic}
\usepackage{amssymb} 
\usepackage{newfloat}
\usepackage{listings}
\DeclareCaptionStyle{ruled}{labelfont=normalfont,labelsep=colon,strut=off} % DO NOT CHANGE THIS
\floatstyle{ruled}
\newfloat{listing}{tb}{lst}{}
\floatname{listing}{Listing}
\usepackage{booktabs}
\usepackage[table]{xcolor}
\usepackage{multirow}

\usepackage{amsmath}
\title{KV-Rescue: Recovering Reasoning Language Model KV Eviction Loss via Stepwise Interleaving}
\author{
    Minsoo Cheong\textsuperscript{\rm 1},
    Woosang Lim\textsuperscript{\rm 1},
    Vincent-Daniel Yun\textsuperscript{\rm 2},
    Sungjoo Yoo\corresponding\textsuperscript{\rm 1}
}
\affiliations{
    \textsuperscript{\rm 1}Seoul National University\\
    \{icycle0409, ftyg656512\}@snu.ac.kr, sungjoo.yoo@gmail.com\\
    \textsuperscript{\rm 2}University of Southern California\\
    yunjuyou@usc.edu\\
}
\begin{document}
\maketitle
\begin{abstract}
KV-cache eviction caps the memory cost of long reasoning traces but is inherently lossy because the model decodes from a partial view of its history. Under aggressive budgets, this not only lowers accuracy but can also cause runaway degeneration, where the model produces incoherent or repetitive tokens until reaching the length limit. We characterize much of this loss as an \emph{information gap} caused by missing context, rather than a \emph{capability gap} caused by limited model capacity. An evicted 7B model and a full-context 1.5B model make complementary errors, and an oracle choice between their answers recovers 79\% of the accuracy gap to the full-KV 7B model. Based on this observation, we propose \textbf{KV-Rescue}, a training-free inference framework that bridges the information gap introduced by KV eviction using a lightweight full-context helper. KV-Rescue interleaves reasoning steps from the two models into a shared trajectory. An online detector uses entropy and compressibility to terminate the generation of incoherent or repetitive base-model candidates early. Across five math benchmarks with Qwen2.5-Math 7B and 72B, KV-Rescue recovers an average of 87\% of the accuracy lost to eviction at eviction budget \(B=64\). A decode-cost analysis further shows that preventing runaway degeneration cuts base-model token generation by 43\% on average.
\end{abstract}
% 코드/데이터셋 링크가 필요하면 abstract 와 본문 사이에 둡니다.
% (익명 제출 시 신원이 드러나지 않게 주의)
% \begin{links}
%     \link{Code}{https://...}
%     \link{Datasets}{https://...}
% \end{links}

% =======================================================================
\section{Introduction}

The growing use of long-context reasoning and agentic systems makes the
KV cache a major bottleneck in LLM inference. KV-cache compression methods address this bottleneck, including quantization~\cite{hooper2024kvquant,liu2024kivi, son2026nsnquant} and merging~\cite{wang2024model}; among them, KV
eviction~\cite{zhang2023h2o, li2024snapkv} is one of the most actively
studied. Recent work tailors eviction to reasoning models by exploiting
their structure~\cite{cai2025rkv} and reports that aggressive eviction can
induce degeneration: the model falls into repetition loops or, in severe
cases, emits incoherent tokens until reaching the length limit. Eviction is
still inherently lossy, since the model sees only a selected part of the
context rather than its full history, and the problem grows more acute at
low budgets.

\begin{figure}[t]
    \centering
    \includegraphics[width=\linewidth]{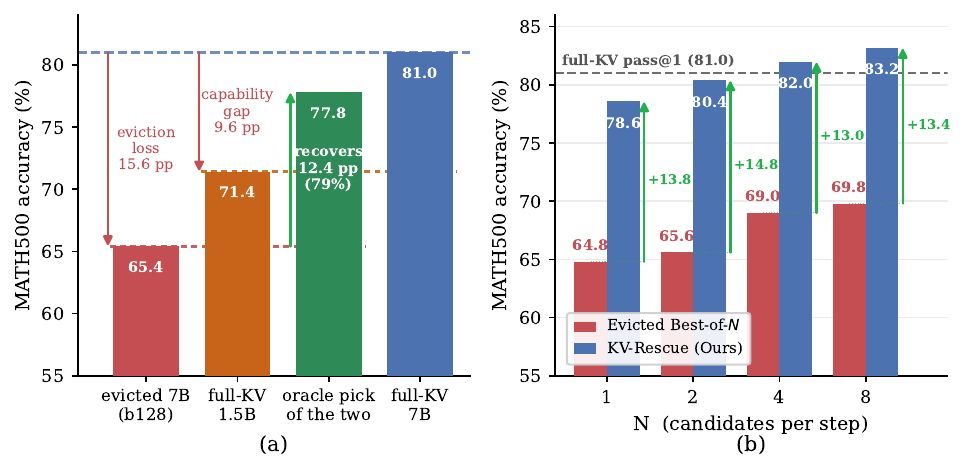}
    \caption{\textbf{(a)} An evicted 7B and a full-KV 1.5B make
    complementary errors; an oracle pick between them recovers $79\%$ of
    the eviction loss. \textbf{(b)} Best-of-$N$ plateaus below full-KV,
    while KV-Rescue (ours) recovers most of the gap and wins at every $N$.
    (MATH500, Qwen2.5-Math-7B, budget~128.)}
    \label{fig:motivation}
\end{figure}

\paragraph{Eviction loss is an information gap, not a capability gap.}
The accuracy lost to eviction is an \emph{information gap}: the model
cannot see parts of its past context. A small model underperforming a
large one instead reflects a \emph{capability gap}: it sees the same
information but has less capacity to solve the problem. These gaps arise
from different limitations, so a small full-context model and an evicted
large model can make complementary errors. Figure~\ref{fig:motivation}(a)
supports this view: an oracle choice between an evicted 7B model and a
full-context 1.5B model recovers $79\%$ of the accuracy gap to the full-KV
7B model.

\paragraph{KV Eviction and Test-Time Scaling.}
It remains unclear whether test-time scaling alone can recover the loss
caused by eviction. We first apply stepwise best-of-$N$ to the evicted
model. As Figure~\ref{fig:motivation}(b) shows, increasing $N$ improves
accuracy but eventually plateaus below full-KV. All candidates are sampled
from the same partial cache state, so additional sampling does not restore
the missing context.

\medskip

Motivated by these observations, we propose \textbf{KV-Rescue}, a
training-free inference framework that bridges the information gap
introduced by KV eviction with a lightweight full-context helper. KV-Rescue
interleaves reasoning steps from the evicted base model and the helper. At
each step, both models generate candidates from their respective cache
states, and a process reward model selects the step that extends the shared
trajectory. We generalize this procedure to a best-of-$N$ setting that draws
several candidates from each model. On MATH500 at budget~128
(Figure~\ref{fig:motivation}(b)), KV-Rescue surpasses
full-KV pass@1 at $N{=}4$ and continues to improve as $N$ increases.
Across five math benchmarks with Qwen2.5-Math 7B and 72B, KV-Rescue
recovers an average of $87\%$ of the accuracy lost to eviction at eviction budget \(B=64\). It also suppresses runaway degeneration; in
our decode-cost analysis, this reduces base-model token generation by
$43\%$ on average.

Our contributions are as follows:
\begin{itemize}
\item We identify KV eviction loss as an \emph{information gap} distinct
from a model's \emph{capability gap}, and show that an evicted large model
and a small full-context model make complementary errors.
\item We introduce \textbf{KV-Rescue}, a training-free inference framework
for context-complementary reasoning between an evicted base model and a
lightweight full-context helper, with an online detector that terminates
degenerate base candidates during generation.
\item We evaluate KV-Rescue across five math benchmarks, two base-model
sizes, and two eviction methods, showing that its recovery persists across
model scales and eviction policies while controlling runaway degeneration.
\end{itemize}
% =======================================================================

\section{Preliminaries}

\subsection{KV Eviction}

At decoding step $t$, a Transformer caches the key--value pairs of all
preceding tokens,
$\mathcal{C}_t=\{(k_i,v_i)\}_{i<t}$, where we suppress layer and attention-head
indices for simplicity. The cache grows linearly with the sequence length and
becomes a major memory cost for long reasoning traces. KV eviction limits the
number of retained entries to a budget $B$ by assigning each entry an
importance score $s_i$, such as accumulated attention, and retaining the
highest-scoring entries~\citep{cai2025rkv,li2024snapkv,zhang2023h2o}:
\begin{equation}
\begin{aligned}
\mathcal{I}_t^B
    &= \operatorname{TopB}\!\left(\{s_i\}_{i<t}\right),\\
\mathcal{C}_t^B
    &= \{(k_i,v_i) : i\in\mathcal{I}_t^B\},\\
\tilde{a}_t
    &= \mathrm{Attn}\!\left(q_t,\mathcal{C}_t^B\right),
\end{aligned}
\end{equation}
where $\operatorname{TopB}$ returns the indices of the $B$ largest scores.
Attention is computed over $\mathcal{C}_t^B$ rather than the full
cache $\mathcal{C}_t$. Once an entry is evicted, subsequent decoding steps
cannot access it without recomputation or retrieval. Reducing $B$ thus
restricts the historical information available to future predictions.

\subsection{PRM-Guided Stepwise Best-of-$N$}

Best-of-$N$~\citep{cobbe2021training} samples $N$ candidates and selects the
one assigned the highest score. Stepwise best-of-$N$~\citep{snell2024scaling}
applies this selection at each reasoning step rather than over complete
trajectories, allowing the scorer to guide generation as it unfolds.

A solution to a question $q$ is represented as a sequence of reasoning steps
$\mathbf{s}=(s_1,\ldots,s_T)$. At step $t$, the committed context is
$c_t=(q,s_{<t})$. A process reward model
(PRM)~\citep{uesato2022solving,lightman2024let,wang2024math,qwen25math}
assigns a score $R(c_t,s)\in\mathbb{R}$ to a candidate step $s$. Given a
policy $\pi$, stepwise best-of-$N$ samples and selects a step as
\begin{equation}
\begin{aligned}
s_t^{(j)}
  &\sim \pi\!\left(\cdot \mid c_t\right),
  \quad j=1,\ldots,N,\\
j_t^\star
  &= \operatorname*{arg\,max}_{j\in\{1,\ldots,N\}}
     R\!\left(c_t,s_t^{(j)}\right),\\
s_t
  &= s_t^{(j_t^\star)}.
\end{aligned}
\end{equation}
The context is then updated to $c_{t+1}=(q,s_{\le t})$, and decoding
continues. Because the candidates share the same committed prefix, they can
be scored in a single batched forward pass.

\subsection{Reasoning Model Degeneration}
\label{sec:prelim_degen}

\begin{figure}[t]
    \centering
    \includegraphics[width=\linewidth]{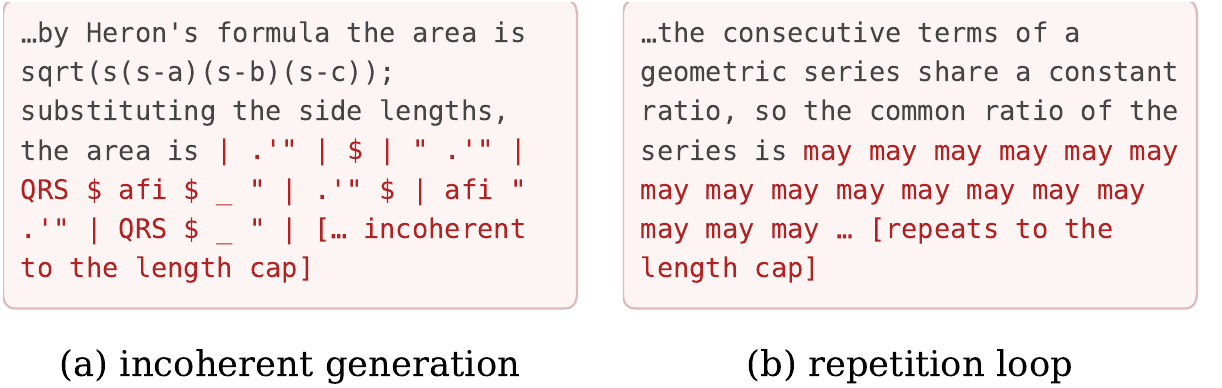}
    \caption{Two degeneration modes observed under aggressive KV eviction in
    Qwen2.5-Math-7B stepwise best-of-$N$ decoding traces. Coherent reasoning
    (gray) transitions into (a) incoherent generation or (b) a repetition
    loop, and continues to the length limit without producing an answer.}
    \label{fig:degen_types}
\end{figure}

Autoregressive generation can degenerate into off-task or repetitive text
instead of completing the response~\citep{xu2022learning}. Recent work
reports that aggressive KV eviction can make such failures more frequent in
reasoning models~\citep{cai2025rkv}. We observe two recurring modes in our
setting (Figure~\ref{fig:degen_types}). \textbf{Incoherent generation}
produces high-entropy sequences of unrelated punctuation, symbols, and
mixed-script tokens. \textbf{Repetition loops} repeatedly emit the same token
or phrase until reaching the length limit. Both modes reduce answer accuracy
and waste decoding computation. We target these two discrete failure modes;
more gradual failures, such as unnecessarily prolonged but coherent
reasoning~\citep{chen2024overthinking,lotfi2026quantized}, are outside the
scope of this work.

 \begin{figure*}[t] % [htbp]는 이미지가 위치할 우선순위입니다.
    \centering % 이미지를 가운데 정렬
    \includegraphics[width=0.9\textwidth]{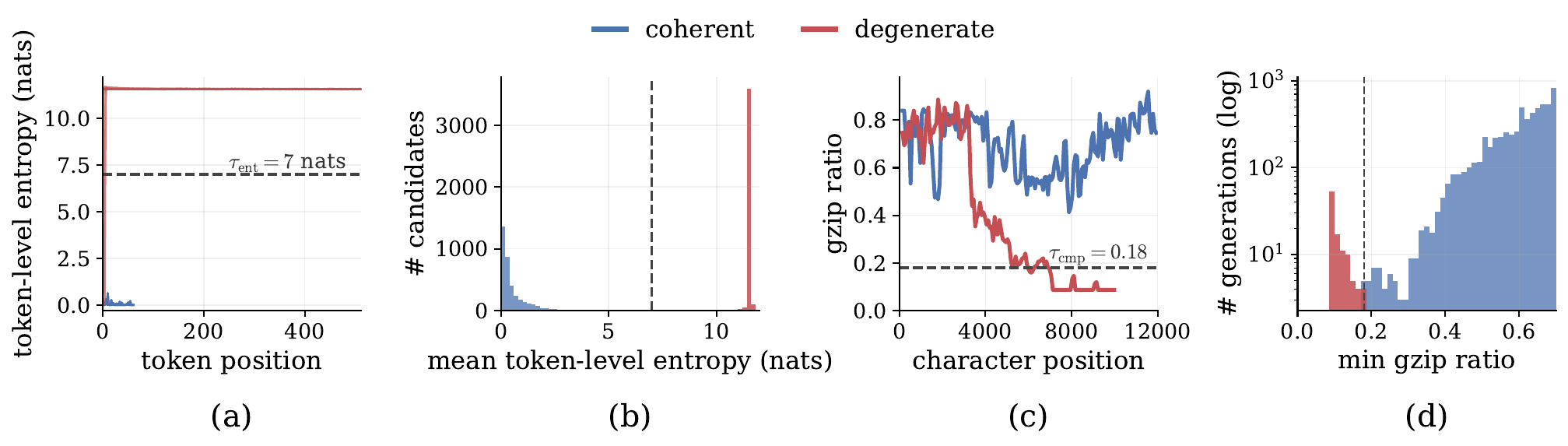}
    \caption{Online degeneration detection (Qwen2.5-Math-7B, R-KV eviction). Both signals are computed from statistics already available during decoding. \textbf{(a,b)} Incoherent generation has high token-level entropy, so coherent and incoherent
    candidates split into two modes that a flat $\tau_{\mathrm{ent}}$ separates without labels.
    \textbf{(c,d)} Repetition loops are confident (low entropy) but highly redundant: the gzip ratio (compressed
    size of a sliding text window over its raw size; lower means more repetitive) collapses below
    $\tau_{\mathrm{cmp}}$, leaving a small loop population cleanly below the coherent bulk.}
    \label{fig:online_detection} % 본문에서 인용할 때 쓸 라벨
\end{figure*}
 
\begin{figure*}[t]
  \centering
  \includegraphics[width=0.7\textwidth]{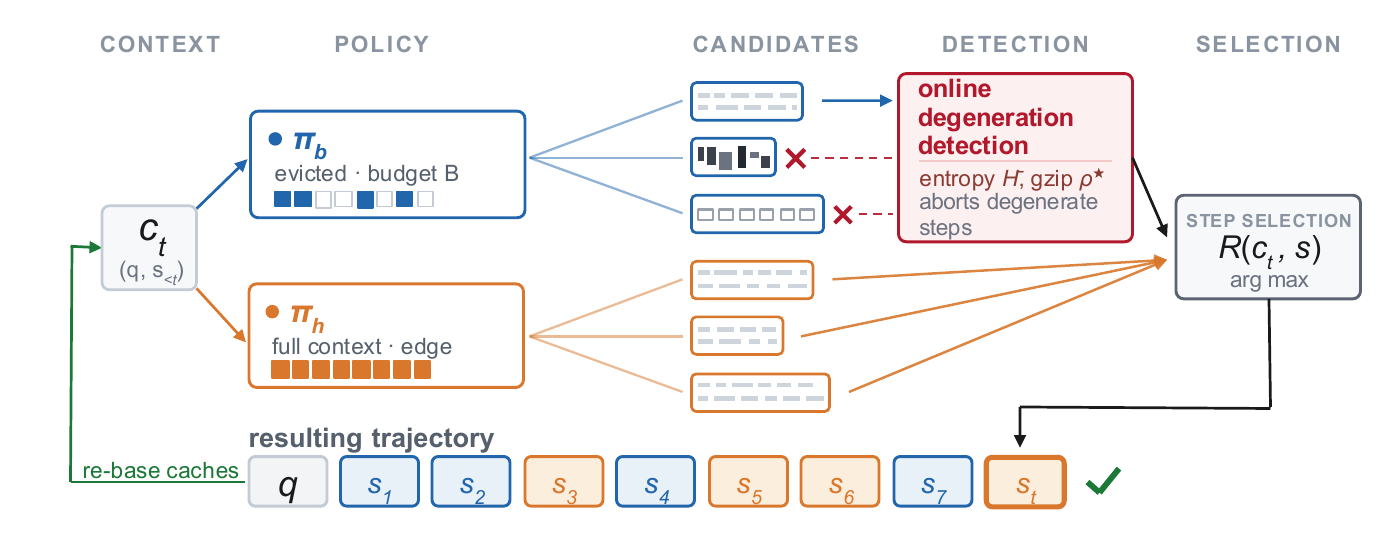}
  \caption{\textbf{KV-Rescue.} At each step, candidates are sampled from an
  evicted base $\pi_b$ and a full-context helper $\pi_h$; an online degeneration
  detector prunes degenerate base steps, the PRM commits the best step $s_t$,
  and base and helper steps interleave. The best-of-$N$ baseline for comparison
  is in the Appendix}
  \label{fig:method}
\end{figure*}

\section{Method}

\subsection{Stepwise Interleaved Reasoning}
\label{sec:interleave}

KV-Rescue extends the answer-level complementarity in
Figure~\ref{fig:motivation} to stepwise reasoning. At each step, it selects
between candidates from a KV-evicted base model and a lightweight
full-context helper using a process reward model (PRM).

\paragraph{Setup.}
Given a question $q$, a solution is a sequence of steps
$\mathbf{s}=(s_1,\dots,s_T)$ delimited by
``\texttt{\textbackslash n\textbackslash n}''. We write $c_t=(q,s_{<t})$
for the committed context and $R(c_t,s)\in[0,1]$ for the PRM score of a
candidate $s$. The base model $\pi_b$ conditions on an evicted cache state
$\mathcal{E}_B(c_t)$ with budget $B$, while the helper $\pi_h$ conditions
on the full context $c_t$. The models thus share the same trajectory but
access different views of its history.

\paragraph{Stepwise interleaving.}
At step $t$, we sample $N_b$ base candidates and $N_h$ helper candidates,
then commit the candidate with the highest PRM score:
\begin{equation}
\label{eq:bonn}
\begin{aligned}
s_t^{b,i}
  &\sim \pi_b\!\left(\cdot \mid \mathcal{E}_B(c_t)\right),
  \quad i=1,\ldots,N_b,\\
s_t^{h,j}
  &\sim \pi_h\!\left(\cdot \mid c_t\right),
  \quad j=1,\ldots,N_h,\\
\mathcal C_t^b
  &= \{s_t^{b,i}\}_{i=1}^{N_b},
  \qquad
  \mathcal C_t^h = \{s_t^{h,j}\}_{j=1}^{N_h},\\
s_t
  &= \operatorname*{arg\,max}_{
       s\in\mathcal C_t^b\cup\mathcal C_t^h
     } R(c_t,s).
\end{aligned}
\end{equation}
The context is then extended to $c_{t+1}=(c_t,s_t)$. If a base candidate is
selected, its fork becomes the next persistent base state. If a helper
candidate is selected, the base model processes the selected tokens and
then applies eviction. At the next step, the helper again receives the full
committed context. Both models continue from the same trajectory
while preserving their distinct context views.

\paragraph{Best-of-$(N_b{+}N_h)$.}
The one-candidate case uses $N_b{=}N_h{=}1$. Our experiments use
$N_b{=}N_h{=}N$, denoted best-of-$N{+}N$, and compare against
eviction-only best-of-$N$ with $N_b{=}N$ and $N_h{=}0$. This matches the
base-model candidate count and eviction budget; the helper cost is analyzed in the efficiency analysis below.

\subsection{Online Degeneration Detection}
\label{sec:detect}

Under aggressive eviction, base candidates exhibit two recurring failure
modes: incoherent generation and repetition loops. Figure
\ref{fig:online_detection} shows that they have distinct online signals,
which we use to terminate degenerate base candidates before PRM scoring.

\paragraph{Incoherent generation.}
Incoherent candidates exhibit sustained high predictive entropy. At
generated position $e$, we compute the entropy and its moving average over
a recent window $W_e$:
\begin{equation}
H_e=-\!\sum_v p_e(v)\log p_e(v), \qquad
\bar H_e=\frac{1}{|W_e|}\sum_{r\in W_e}H_r.
\end{equation}
A candidate is terminated when
$\bar H_e>\tau_{\mathrm{ent}}$.

\paragraph{Repetition loops.}
Loops are typically low-entropy but highly compressible. Let $z_e$ denote
a fixed-width window over the most recent candidate text. Once the window
is available, we compute
\begin{equation}
\rho_e=\frac{|\mathrm{gzip}(z_e)|}{|z_e|}
\end{equation}
and terminate the candidate when
$\rho_e<\tau_{\mathrm{cmp}}$. Figure
\ref{fig:online_detection}(d) reports the per-candidate minimum
$\rho^\star(s)=\min_e\rho_e$ for analysis, while online detection uses the
current $\rho_e$.

The detector fires at the first eligible position satisfying either test:
\begin{equation}
\mathrm{deg}_e(s)=
\mathbb{1}[\bar H_e>\tau_{\mathrm{ent}}]
\vee
\mathbb{1}[\rho_e<\tau_{\mathrm{cmp}}].
\end{equation}
Only base candidates are screened. With
$\mathrm{deg}(s)=\bigvee_e\mathrm{deg}_e(s)$, selection becomes
\begin{equation}
\begin{aligned}
\widehat{\mathcal C}_t
&=\{s\in\mathcal C_t^b:\neg\mathrm{deg}(s)\}\cup\mathcal C_t^h,\\
s_t&=\operatorname*{arg\,max}_{s\in\widehat{\mathcal C}_t}R(c_t,s).
\end{aligned}
\end{equation}
Helper candidates keep the pool nonempty even if all base candidates are
aborted. Both signals use quantities already available during decoding and
require no additional model forward pass. 
\subsection{KV-Rescue}
\label{sec:kvrescue}

KV-Rescue combines stepwise interleaving with online degeneration detection
(Figure~\ref{fig:method}), committing the highest-scoring candidate among
the surviving base candidates and the full-context helper candidates.

% =======================================================================

\begin{table}[t]
\centering
\caption{Decode cost of best-of-$N$ (BoN) vs.\ KV-Rescue (KVR) on Qwen2.5-Math-7B ($N{=}8$): per-problem mean
steps, output length, and generated 7B-base / 1.5B-small tokens. Cells are heat-shaded per dataset and metric
(green = low, red = high); as the KV budget shrinks the evicted-only BoN runs away (red) while KVR stays
controlled (green). \textit{full}~=~full-KV;}
\label{tab:gen_length}
\resizebox{\linewidth}{!}{% Qwen2.5-Math-7B (N=8) decode efficiency under R-KV eviction. bon=best-of-N, b2n=best-of-2N (ours).
% Per-problem means; cells heat-shaded per (dataset,metric). full = full-KV (bon only).
% Walltime/PRM excluded. Needs \usepackage[table]{xcolor} and \usepackage{booktabs}.
\begin{tabular}{ll rr rr rr r}
\toprule
& & \multicolumn{2}{c}{Steps} & \multicolumn{2}{c}{Out.\ len} & \multicolumn{2}{c}{Base 7B tok} & 1.5B tok \\
\cmidrule(lr){3-4}\cmidrule(lr){5-6}\cmidrule(lr){7-8}\cmidrule(lr){9-9}
Dataset & Bud. & BoN & KVR & BoN & KVR & BoN & KVR & KVR \\
\midrule
MATH500 & \textit{full} & \cellcolor[rgb]{0.877,0.930,0.816} 27.0 & -- & \cellcolor[rgb]{0.859,0.925,0.828} 673 & -- & \cellcolor[rgb]{0.857,0.924,0.829} 6,106 & -- & -- \\
 & 512 & \cellcolor[rgb]{0.891,0.934,0.806} 31.8 & \cellcolor[rgb]{0.842,0.921,0.839} 15.4 & \cellcolor[rgb]{0.856,0.924,0.829} 670 & \cellcolor[rgb]{0.872,0.929,0.818} 687 & \cellcolor[rgb]{0.862,0.926,0.825} 6,189 & \cellcolor[rgb]{0.840,0.920,0.840} 5,841 & 6,102 \\
 & 256 & \cellcolor[rgb]{0.936,0.946,0.776} 46.7 & \cellcolor[rgb]{0.840,0.920,0.840} 14.8 & \cellcolor[rgb]{0.887,0.932,0.809} 701 & \cellcolor[rgb]{0.840,0.920,0.840} 653 & \cellcolor[rgb]{0.895,0.935,0.804} 6,704 & \cellcolor[rgb]{0.841,0.920,0.839} 5,855 & 5,931 \\
 & 128 & \cellcolor[rgb]{0.987,0.959,0.742} 63.8 & \cellcolor[rgb]{0.842,0.921,0.838} 15.6 & \cellcolor[rgb]{0.939,0.946,0.774} 754 & \cellcolor[rgb]{0.852,0.923,0.832} 666 & \cellcolor[rgb]{0.952,0.950,0.766} 7,607 & \cellcolor[rgb]{0.854,0.924,0.830} 6,068 & 6,142 \\
 & 64 & \cellcolor[rgb]{0.840,0.400,0.380} 114 & \cellcolor[rgb]{0.859,0.925,0.827} 21.1 & \cellcolor[rgb]{0.840,0.400,0.380} 960 & \cellcolor[rgb]{0.857,0.925,0.829} 671 & \cellcolor[rgb]{0.840,0.400,0.380} 10,587 & \cellcolor[rgb]{0.871,0.928,0.820} 6,327 & 6,560 \\
\midrule
AIME24 & \textit{full} & \cellcolor[rgb]{0.974,0.899,0.701} 171 & -- & \cellcolor[rgb]{0.968,0.878,0.687} 1,566 & -- & \cellcolor[rgb]{0.987,0.948,0.732} 16,803 & -- & -- \\
 & 512 & \cellcolor[rgb]{0.977,0.956,0.749} 148 & \cellcolor[rgb]{0.842,0.921,0.838} 43.8 & \cellcolor[rgb]{0.976,0.956,0.749} 1,510 & \cellcolor[rgb]{0.887,0.933,0.809} 1,369 & \cellcolor[rgb]{0.980,0.957,0.747} 16,457 & \cellcolor[rgb]{0.840,0.920,0.840} 12,775 & 13,591 \\
 & 256 & \cellcolor[rgb]{0.907,0.652,0.542} 222 & \cellcolor[rgb]{0.840,0.920,0.840} 41.9 & \cellcolor[rgb]{0.907,0.649,0.540} 1,663 & \cellcolor[rgb]{0.840,0.920,0.840} 1,295 & \cellcolor[rgb]{0.905,0.644,0.537} 18,943 & \cellcolor[rgb]{0.843,0.921,0.838} 12,842 & 13,620 \\
 & 128 & \cellcolor[rgb]{0.840,0.400,0.380} 275 & \cellcolor[rgb]{0.852,0.923,0.832} 51.1 & \cellcolor[rgb]{0.840,0.400,0.380} 1,768 & \cellcolor[rgb]{0.882,0.931,0.812} 1,362 & \cellcolor[rgb]{0.840,0.400,0.380} 20,660 & \cellcolor[rgb]{0.855,0.924,0.830} 13,159 & 14,307 \\
 & 64 & \cellcolor[rgb]{0.899,0.622,0.523} 229 & \cellcolor[rgb]{0.870,0.928,0.820} 64.9 & \cellcolor[rgb]{0.947,0.798,0.636} 1,600 & \cellcolor[rgb]{0.899,0.936,0.801} 1,388 & \cellcolor[rgb]{0.843,0.412,0.388} 20,573 & \cellcolor[rgb]{0.871,0.928,0.819} 13,584 & 15,812 \\
\midrule
AMC23 & \textit{full} & \cellcolor[rgb]{0.879,0.931,0.814} 40.4 & -- & \cellcolor[rgb]{0.859,0.925,0.828} 966 & -- & \cellcolor[rgb]{0.851,0.923,0.832} 9,377 & -- & -- \\
 & 512 & \cellcolor[rgb]{0.867,0.927,0.822} 35.2 & \cellcolor[rgb]{0.844,0.921,0.838} 25.0 & \cellcolor[rgb]{0.853,0.923,0.831} 959 & \cellcolor[rgb]{0.897,0.935,0.802} 1,011 & \cellcolor[rgb]{0.868,0.927,0.821} 9,695 & \cellcolor[rgb]{0.862,0.926,0.826} 9,574 & 9,476 \\
 & 256 & \cellcolor[rgb]{0.957,0.951,0.762} 73.6 & \cellcolor[rgb]{0.840,0.920,0.840} 23.4 & \cellcolor[rgb]{0.888,0.933,0.808} 1,000 & \cellcolor[rgb]{0.878,0.930,0.815} 988 & \cellcolor[rgb]{0.906,0.938,0.796} 10,440 & \cellcolor[rgb]{0.840,0.920,0.840} 9,159 & 10,350 \\
 & 128 & \cellcolor[rgb]{0.914,0.675,0.557} 121 & \cellcolor[rgb]{0.857,0.925,0.829} 30.7 & \cellcolor[rgb]{0.951,0.813,0.646} 1,166 & \cellcolor[rgb]{0.901,0.936,0.800} 1,015 & \cellcolor[rgb]{0.956,0.834,0.659} 12,706 & \cellcolor[rgb]{0.866,0.927,0.823} 9,654 & 10,451 \\
 & 64 & \cellcolor[rgb]{0.840,0.400,0.380} 152 & \cellcolor[rgb]{0.860,0.925,0.827} 32.1 & \cellcolor[rgb]{0.840,0.400,0.380} 1,295 & \cellcolor[rgb]{0.840,0.920,0.840} 944 & \cellcolor[rgb]{0.840,0.400,0.380} 14,947 & \cellcolor[rgb]{0.843,0.921,0.838} 9,209 & 9,978 \\
\midrule
GSM8K & \textit{full} & \cellcolor[rgb]{0.846,0.922,0.836} 8.3 & -- & \cellcolor[rgb]{0.844,0.921,0.837} 307 & -- & \cellcolor[rgb]{0.842,0.921,0.839} 2,477 & -- & -- \\
 & 512 & \cellcolor[rgb]{0.847,0.922,0.835} 8.8 & \cellcolor[rgb]{0.840,0.920,0.840} 6.6 & \cellcolor[rgb]{0.845,0.921,0.837} 307 & \cellcolor[rgb]{0.840,0.920,0.840} 303 & \cellcolor[rgb]{0.843,0.921,0.838} 2,500 & \cellcolor[rgb]{0.840,0.920,0.840} 2,448 & 2,317 \\
 & 256 & \cellcolor[rgb]{0.851,0.923,0.833} 9.9 & \cellcolor[rgb]{0.841,0.920,0.840} 6.8 & \cellcolor[rgb]{0.851,0.923,0.833} 314 & \cellcolor[rgb]{0.844,0.921,0.838} 306 & \cellcolor[rgb]{0.848,0.922,0.834} 2,577 & \cellcolor[rgb]{0.842,0.921,0.838} 2,484 & 2,380 \\
 & 128 & \cellcolor[rgb]{0.915,0.940,0.790} 28.9 & \cellcolor[rgb]{0.841,0.920,0.839} 7.0 & \cellcolor[rgb]{0.909,0.939,0.794} 376 & \cellcolor[rgb]{0.844,0.921,0.837} 307 & \cellcolor[rgb]{0.903,0.937,0.798} 3,422 & \cellcolor[rgb]{0.846,0.922,0.836} 2,537 & 2,446 \\
 & 64 & \cellcolor[rgb]{0.840,0.400,0.380} 96.2 & \cellcolor[rgb]{0.846,0.922,0.836} 8.5 & \cellcolor[rgb]{0.840,0.400,0.380} 619 & \cellcolor[rgb]{0.846,0.922,0.836} 309 & \cellcolor[rgb]{0.840,0.400,0.380} 7,056 & \cellcolor[rgb]{0.850,0.923,0.833} 2,607 & 2,590 \\
\midrule
OlympiadBench & \textit{full} & \cellcolor[rgb]{0.897,0.935,0.802} 57.5 & -- & \cellcolor[rgb]{0.840,0.920,0.840} 1,016 & -- & \cellcolor[rgb]{0.842,0.920,0.839} 9,499 & -- & -- \\
 & 512 & \cellcolor[rgb]{0.949,0.949,0.767} 84.9 & \cellcolor[rgb]{0.840,0.920,0.840} 26.8 & \cellcolor[rgb]{0.896,0.935,0.803} 1,092 & \cellcolor[rgb]{0.864,0.926,0.824} 1,049 & \cellcolor[rgb]{0.900,0.936,0.800} 11,040 & \cellcolor[rgb]{0.844,0.921,0.838} 9,550 & 10,189 \\
 & 256 & \cellcolor[rgb]{0.975,0.956,0.750} 98.8 & \cellcolor[rgb]{0.844,0.921,0.837} 29.1 & \cellcolor[rgb]{0.911,0.939,0.793} 1,113 & \cellcolor[rgb]{0.847,0.922,0.835} 1,026 & \cellcolor[rgb]{0.924,0.943,0.784} 11,673 & \cellcolor[rgb]{0.840,0.920,0.840} 9,458 & 10,242 \\
 & 128 & \cellcolor[rgb]{0.929,0.733,0.594} 139 & \cellcolor[rgb]{0.858,0.925,0.828} 36.2 & \cellcolor[rgb]{0.981,0.927,0.719} 1,233 & \cellcolor[rgb]{0.867,0.927,0.822} 1,053 & \cellcolor[rgb]{0.975,0.903,0.703} 13,793 & \cellcolor[rgb]{0.854,0.924,0.831} 9,819 & 13,132 \\
 & 64 & \cellcolor[rgb]{0.840,0.400,0.380} 187 & \cellcolor[rgb]{0.858,0.925,0.828} 36.4 & \cellcolor[rgb]{0.840,0.400,0.380} 1,425 & \cellcolor[rgb]{0.857,0.925,0.828} 1,040 & \cellcolor[rgb]{0.840,0.400,0.380} 17,327 & \cellcolor[rgb]{0.867,0.927,0.822} 10,171 & 10,920 \\
\bottomrule
\end{tabular}
}
\end{table}

\begin{figure*}[t] % [htbp]는 이미지가 위치할 우선순위입니다.
    \centering % 이미지를 가운데 정렬
    \includegraphics[width=0.9\textwidth]{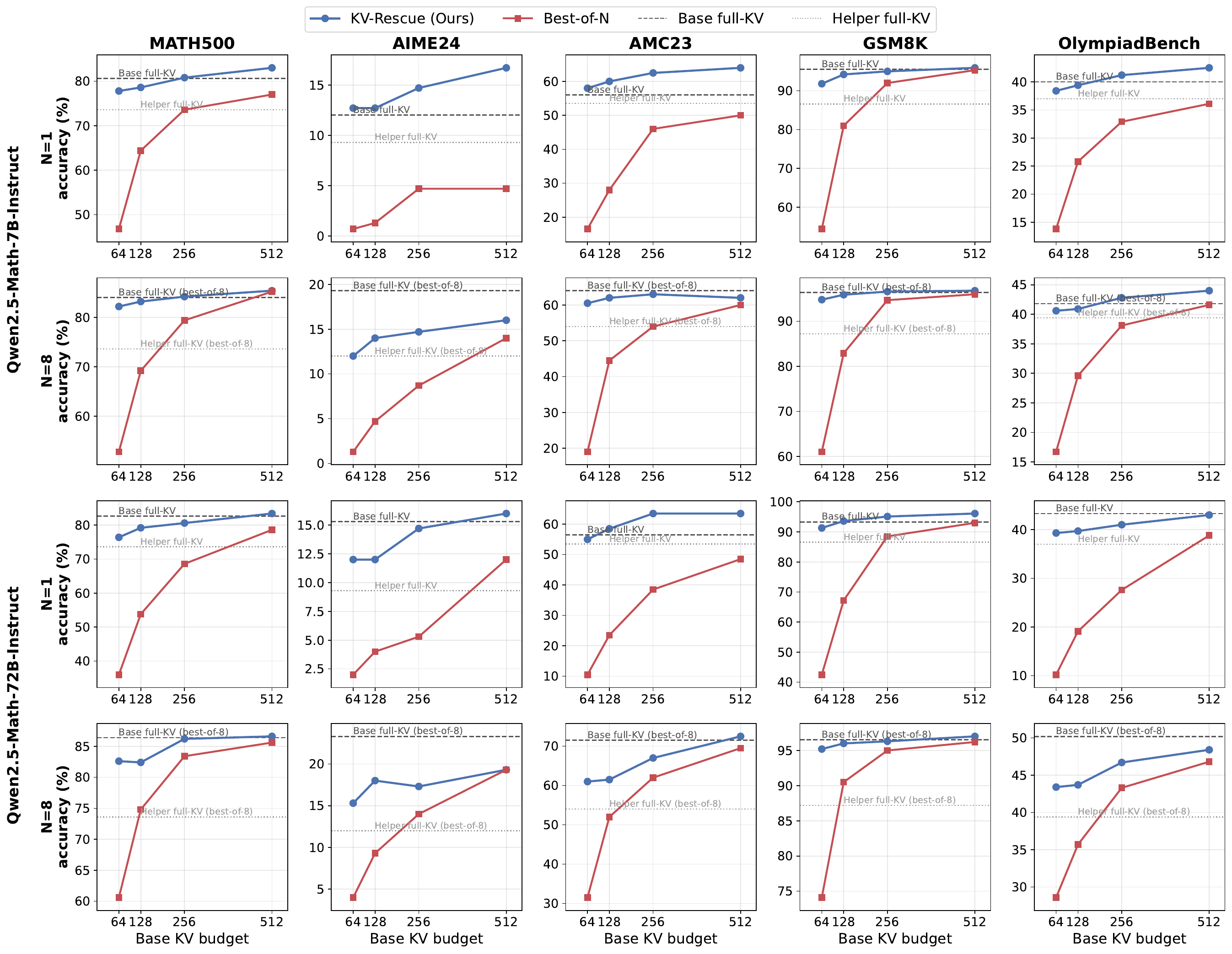}
    \caption{\textbf{Main results.} pass@1 vs.\ base KV budget across the $4\times5$
grid (Qwen2.5-Math 7B/72B $\times$ $N{=}1,8$; MATH500, AIME24, AMC23, GSM8K,
OlympiadBench). Blue is KV-Rescue (ours), red is eviction-only best-of-$N$;
dashed and dotted lines mark the base full-KV and helper full-KV references.
Best-of-$N$ falls off sharply as the budget shrinks, while KV-Rescue stays close
to base full-KV across budgets.}
    \label{fig:main_results} % 본문에서 인용할 때 쓸 라벨
\end{figure*}

\section{Experiments}

\subsection{Experiment Setup}
\label{sec:setup}

\paragraph{Models.}
We use the Qwen2.5-Math family~\cite{qwen25math}, which provides reasoning
models at multiple scales together with a matching PRM, reducing distribution
mismatch in step scoring~\cite{khalifa2025process,zhu2025retrieval}. The base
is \texttt{Qwen2.5-Math-7B-Instruct} or
\texttt{Qwen2.5-Math-72B-Instruct} (hereafter \emph{7B} and \emph{72B}),
the full-context helper is \texttt{Qwen2.5-Math-1.5B-Instruct}, and the PRM
is \texttt{Qwen2.5-Math-PRM-7B}. We retain the 7B PRM for the 72B base to
test whether a smaller verifier can guide the larger model. All models run
in \texttt{bfloat16}, with context capped at $4096$ tokens.

\paragraph{Benchmarks and decoding.}
We report pass@1 on MATH500~\cite{hendrycks2021math},
AIME~2024~\cite{aime2024}, AMC~2023~\cite{mathai_amc23},
GSM8K~\cite{cobbe2021training}, and
OlympiadBench~\cite{he2024olympiadbench}, using the official Qwen2.5-Math
grader. AIME and AMC are averaged over five seeds, and the remaining
benchmarks use one run. We sample at temperature $0.8$, top-$p$ $0.95$,
and top-$k$ $20$.

Main results use $N\in\{1,8\}$. For both methods, $N$ denotes the number
of candidates drawn from the base model; KV-Rescue additionally draws
$N$ candidates from the 1.5B helper. The comparison matches
the base-model candidate budget and eviction budget, rather than the
total number of policy candidates. The helper requires approximately
24\% and 2.5\% of the per-token decode compute of the 7B and 72B bases, respectively.

\paragraph{KV eviction.}
We use R-KV~\cite{cai2025rkv} with $\lambda=0.1$ and a protected recent
window of 128 tokens. The reported budget $B\in\{64,128,256,512\}$ controls
the score-selected entries, giving a physical cache bound of
$B'=B+128$. We also evaluate a no-eviction full-KV reference.

\paragraph{Implementation and hardware.}
We build on the v1 engine of a modified \texttt{vLLM}
\cite{kwon2023efficient}. Efficiency and latency are measured on A100 SXM
GPUs with 40GB memory. The appendix provides full implementation details
and repeats the 7B evaluation with SnapKV~\cite{li2024snapkv}, showing that
the recovery is not specific to R-KV.

\subsection{Experiment Results}

\paragraph{Accuracy versus KV Budget.}
Figure~\ref{fig:main_results} compares accuracy across the $4\times5$ grid.
Eviction-only best-of-$N$ degrades sharply as $B$ decreases and often
approaches or falls below the full-context helper at the smallest budget.
KV-Rescue remains substantially more stable and closer to the full-KV
reference. At $B=64$, it recovers an average of $87\%$ of the full-KV
accuracy loss across the grid, using
$(A_{\mathrm{KVR}}-A_{\mathrm{evict}})/
(A_{\mathrm{full}}-A_{\mathrm{evict}})$ for each setting. Increasing $N$
generally improves both methods but does not eliminate the gap under
aggressive eviction. The same recovery pattern persists with the 72B base
and 7B PRM.

\paragraph{Generation Length.}
Table~\ref{tab:gen_length} compares reasoning-step count, output length,
and base-model token generation. As $B$ decreases, eviction-only
best-of-$N$ exhibits severe length inflation as degenerate generations
approach the length cap. KV-Rescue prevents these runaways and remains
closer to the full-KV reference. At $B=64$ in the 7B, $N=8$ analysis, it
reduces base-model token generation by $43\%$ on average across benchmarks.

% N-scaling (대표 pair Qwen2.5). placement/거동은 모델 독립적 → 한 pair.
% 표로 제시.

% =======================================================================
\section{Discussion}

\subsection{Recovering KV-Eviction-Induced Information Loss}
\label{sec:discuss_recovery}

\begin{figure}[t]
  \centering
  \includegraphics[width=0.9\linewidth]{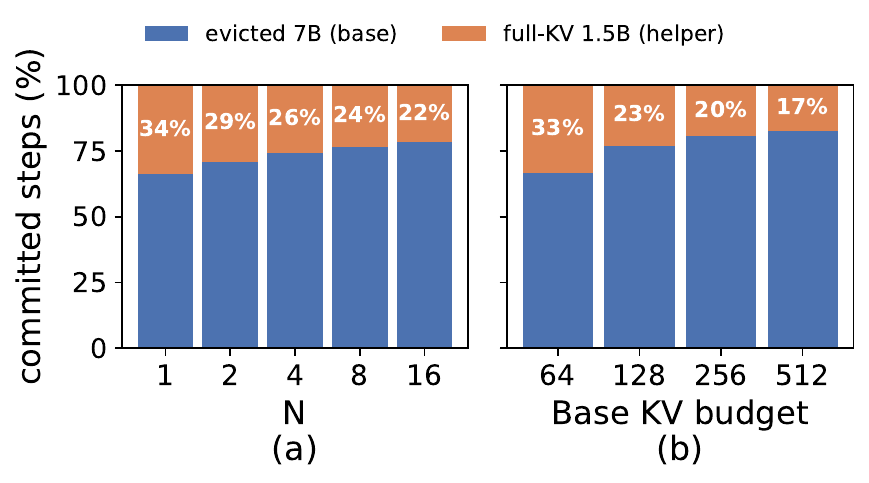}
  \caption{Share of committed steps selected from the evicted 7B base and
  the full-context 1.5B helper on MATH500 with best-of-$N{+}N$ decoding:
  \textbf{(a)} as $N$ varies at budget~128 and \textbf{(b)} as the KV budget
  varies at $N{=}8$.}
  \label{fig:selection_ratio}
\end{figure}

\begin{figure*}[t]
  \centering
  \includegraphics[width=\textwidth]{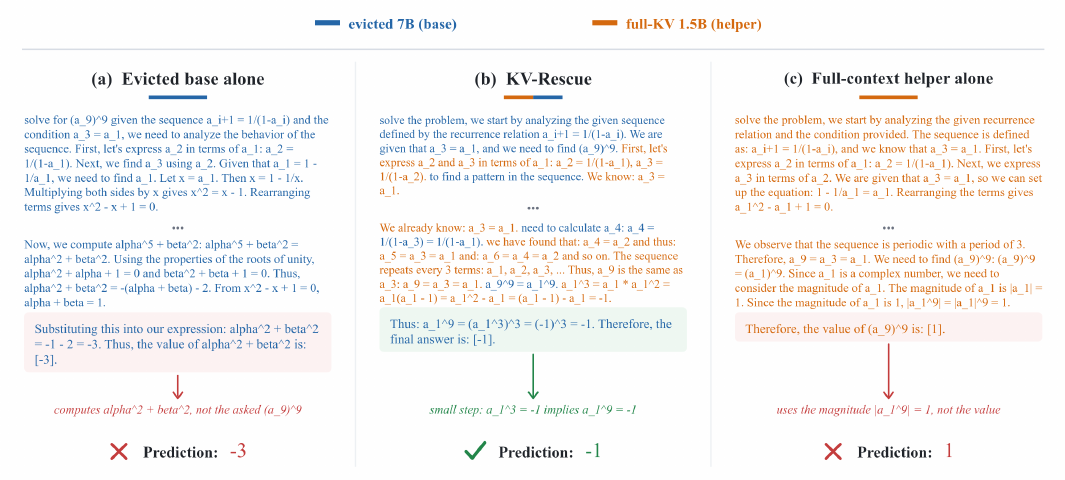}
  \caption{A representative problem solved by KV-Rescue but not by either
  model's independent trajectory: \textbf{(a)} the evicted base,
  \textbf{(b)} KV-Rescue, and \textbf{(c)} the full-context helper. Blue and
  orange mark committed base and helper steps, respectively.}
  \label{fig:selection_example}
\end{figure*}

We examine how the two models contribute to the recovered trajectory.
Figure~\ref{fig:selection_ratio} shows that the helper is not merely a rare
fallback: it supplies $17$--$34\%$ of committed steps across the evaluated
settings. Its share increases from $17\%$ at budget 512 to $33\%$ at budget
64, consistent with full-context candidates becoming more useful as the
base loses more history. The base nevertheless contributes most steps, and
the source ratio changes little as $N$ grows, indicating that KV-Rescue
combines the two models rather than replacing the base with the helper.

Figure~\ref{fig:selection_example} illustrates this complementarity. The
evicted base loses track of the requested quantity, while the helper retains
the target but makes an error in the derivation. By interleaving steps from
the two models, KV-Rescue avoids both failures and reaches the correct answer.
Together, the selection trend and the example support the intended mechanism:
the helper contributes selectively, and its role grows as eviction becomes
more severe.

% (a) best-of-N vs best-of-2N trajectory 예시 비교.
% (b) budget 에 따른 main vs small 기여도 변화 (small-win%).
% TODO: complementarity / beyond-union (둘 다 못 푸는 걸 b2n 이 푼 수) +
%       emergent 케이스 src 구성 확인 → 추후 추가.

% =====================================================================
% PRM verifier-bounded analysis (condensed)
% oracle = per-problem OR of each source's INDEPENDENT trajectory.
% Negative gap (b64 in-domain, -0.6) = b2n exceeds trajectory oracle
%   via per-step source mixing. Replace Appendix~X with real label.
% =====================================================================

\subsection{Degeneration Analysis}
\label{sec:discuss_degen}
\begin{table}[t]
\centering
\caption{Ablation of interleaving and early exit on MATH500
($N{=}8$, Qwen2.5-Math-7B). Rows are cumulative. \emph{Degen.} is the
fraction of incorrect trajectories with at least 256 committed steps, used
to identify context-length runaways. \emph{Base tok.} is the mean number of
tokens generated by base-model candidates per problem; \emph{Steps} is the
mean number of committed steps; and \emph{EE} is the fraction of base
candidates terminated by early exit.}
\label{tab:degen}
\resizebox{\linewidth}{!}{\begin{tabular}{llrrrrr}
\toprule
Budget & Method & Acc. & Degen. & Base tok. & Steps & EE \\
& & (\%) & (\%) & & & (\%) \\
\midrule
\textit{full} & full-KV & 84.0 & 2.4 & 6,106 & 27 & -- \\
\midrule
\multirow{3}{*}{64}
 & best-of-$N$        & 52.8 & 13.4 & 10,587 & 114 & -- \\
 & \;+ interleaving   & 81.8 &  2.8 &  7,898 &  36 & -- \\
 & \;\;+ early exit   & 82.2 &  0.2 &  6,327 &  21 & 8.0 \\
\midrule
\multirow{3}{*}{128}
 & best-of-$N$        & 69.2 & 6.0 & 7,607 & 64 & -- \\
 & \;+ interleaving   & 81.2 & 1.4 & 6,673 & 25 & -- \\
 & \;\;+ early exit   & 83.2 & 0.0 & 6,068 & 16 & 6.4 \\  
\bottomrule
\end{tabular}}
\end{table}

Table~\ref{tab:degen} isolates the contributions of interleaving and early
exit. At budget 64, eviction-only best-of-$N$ raises the degeneration rate
from $2.4\%$ under full-KV to $13.4\%$. Sampling alone does not prevent these
runaways because every candidate is generated from the same evicted cache.

\paragraph{Interleaving substantially reduces trajectory degeneration.}
Adding full-context helper candidates gives the PRM alternatives that do not
share the base model's missing context. At budget 64, interleaving reduces
degeneration from $13.4\%$ to $2.8\%$, lowers the mean trajectory from 114 to
36 steps, and raises accuracy from $52.8\%$ to $81.8\%$. This accounts for
most of the recovery.

\paragraph{Early exit removes candidate-level token waste.}
Interleaving can avoid committing a degenerate base candidate while still
decoding that candidate to the step limit. Early exit terminates such branches
during generation. At budget 64, it fires on $8.0\%$ of base candidates and
reduces base-model token generation by a further $20\%$
($7.9$k${\to}6.3$k) without reducing accuracy. It also lowers degeneration
from $2.8\%$ to $0.2\%$, while eliminating it under the table's criterion at
budget 128.

\begin{figure}[t]
  \centering
  \includegraphics[width=\linewidth]{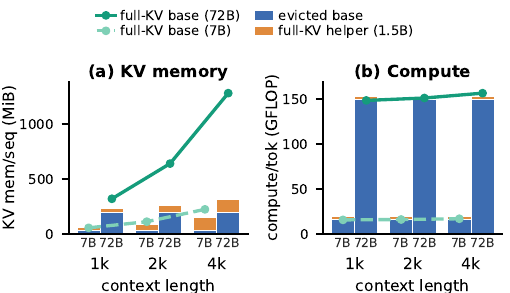}
  \caption{\textbf{Per-sequence KV memory (a) and per-token decode compute
  (b) of KV-Rescue vs.\ context length}, for a 7B and a 72B base (bf16,
  R-KV budget $512$, $B'{=}640$). Bars are KV-Rescue: the \emph{evicted
  base} (blue, bounded by the budget) plus the \emph{full-KV helper} 1.5B
      (orange). Lines are the \emph{full-KV base} baseline.}
  \label{fig:theoretical_overhead}
\end{figure}

\begin{table}[t]
\centering
\caption{\textbf{MATH500 wall-clock efficiency}
(Qwen2.5-Math-72B, $N{=}8$, KV budget 64; optimized engine with batched
eviction and dynamic KV reserve). \emph{prob/min}: steady-state throughput
(excludes one-time load); $t_\text{gen}$: per-problem generation time (base
$\pi_b$ and full-context helper $\pi_h$ run concurrently); $t_\text{score}$:
PRM time; all amortized over batch concurrency. Max length 4096;
full-KV OOMs at batch 64.}
\label{tab:throughput}
\resizebox{\linewidth}{!}{\begin{tabular}{@{}llrrrr@{}}
\toprule
Method & Batch & prob/min & $t_\text{gen}$\,(s) & $t_\text{score}$\,(s) & 72B tok \\
\midrule
full-KV & 8 & 6.7 & 8.41 & 0.375 & 5,392 \\
 & 64 & \multicolumn{4}{c}{\emph{OOM}} \\
\midrule
BoN & 8 & 1.2 & 44.45 & 1.619 & 18,144 \\
 & 64 & 3.2 & 12.84 & 0.193 & 18,371 \\
\midrule
\textbf{KV-Rescue} & 8 & 4.2 & 12.67 & 0.410 & 5,951 \\
 & 64 & 10.6 & 4.58 & 0.034 & 5,460 \\
\bottomrule
\end{tabular}

}
\end{table}

\subsection{Efficiency Analysis}
\label{sec:efficiency}

We analyze the cost of KV-Rescue theoretically and report wall-clock
measurements below; the full derivation is in the appendix.

Figure~\ref{fig:theoretical_overhead} shows KV memory and decode compute
against context length at a fixed generation length, allowing a per-token
comparison. Eviction caps the base model's KV, while the full-KV baseline
grows linearly ($0.31$ vs. $1.25$ GiB at $4$k for the 72B base). The
helper's KV grows with context but remains small, and a full-context 1.5B
forward adds only ${\sim}2.5\%$ of the 72B base's per-token compute
(${\sim}24\%$ for the 7B base). The added model-level cost is
modest, particularly as the base model scales.

This fixed-length comparison isolates the per-token overhead. In practice,
KV-Rescue also generates fewer tokens by preventing the runaway degeneration
that inflates the evicted baseline's output length
(Table~\ref{tab:gen_length}). It combines a lightweight helper
with substantially fewer expensive base-model tokens. The wall-clock
measurements below account for the complete inference pipeline, including
helper generation, PRM scoring, and system overheads.

Table~\ref{tab:throughput} reports MATH500 wall-clock throughput on a single
node of eight A100 40GB GPUs. To keep the total accelerator allocation fixed
across methods, the 72B base is tensor-parallel across all eight GPUs, with
the PRM and 1.5B helper co-located on the same GPU pool. The reported
throughput thus includes resource contention and scheduling overhead from
co-location rather than assuming isolated model execution.

Even under this shared-resource configuration, at batch 64, KV-Rescue
achieves $3.3\times$ the throughput of eviction-only best-of-$N$ at the same
base KV budget, while reducing base-model generation by $70\%$
($18.4k{\to}5.5k$ tokens per problem). Capping the base KV cache also
allows KV-Rescue to serve batch 64, which full-KV cannot fit. PRM scoring
remains a small fraction of generation time, and the helper runs concurrently
with the base. Since the helper has a small compute and memory footprint, it
is also amenable to disaggregated execution on a lower-cost accelerator,
offering a path to further isolate helper execution and improve concurrency.
These results show that KV-Rescue makes aggressive eviction practical by
preventing the degeneration that otherwise inflates its decode cost.

% =======================================================================
\section{Related Work}

\paragraph{KV cache compression.}
A large body of work reduces KV-cache memory through
quantization~\cite{hooper2024kvquant,liu2024kivi,son2026nsnquant},
merging~\cite{wang2024model}, and
eviction~\cite{zhang2023h2o,li2024snapkv}; recent work further adapts
eviction to reasoning models~\cite{cai2025rkv}. These methods operate
within a single model, reducing the precision, redundancy, or number of
entries in its cache. Closest to our setting is
SmallKV~\cite{zhao2026smallkv}, which maintains a full-cache small model
and uses its attention to guide large-model eviction and approximate the
contribution of marginal tokens. The large model nevertheless remains the
sole generator. KV-Rescue instead treats the small full-context model as an
independent reasoning policy and interleaves its steps with those of the
evicted base. It operates above the eviction mechanism and can be orthogonally combined with different eviction policies.

\paragraph{Test-time and multi-model inference.}
Test-time scaling improves reasoning by sampling and selecting candidates
through best-of-$N$~\cite{cobbe2021training} and PRM-guided step or beam
search~\cite{snell2024scaling,lightman2024let}. These methods typically
sample from a single policy conditioned on the same context state. KV-Rescue
instead draws step candidates from two policies with asymmetric context
access, using selection to recover information unavailable to the evicted
base. Speculative decoding similarly combines small and large models, but
uses the small model to draft tokens while preserving the large model's
output distribution~\cite{leviathan2023fast,chen2023accelerating}. In
KV-Rescue, either model may contribute a committed reasoning step, and the
goal is to recover reasoning quality under KV eviction rather than accelerate
an unchanged target policy.
% =======================================================================
\section{Conclusion}

KV eviction bounds the memory cost of long reasoning but can remove
information needed by later steps. We showed that much of the resulting loss
is an information gap rather than a capability gap: an evicted large model
and a lightweight full-context model make complementary errors. Based on
this observation, we introduced \textbf{KV-Rescue}, a training-free inference
framework that interleaves reasoning steps from the two models and terminates
degenerate base candidates online. Across five math benchmarks with
Qwen2.5-Math 7B and 72B, KV-Rescue recovers an average of $87\%$ of the
accuracy lost at an eviction budget of 64, while reducing base-model token
generation by $43\%$ in our decode-cost analysis. At batch 64, KV-Rescue achieves $3.3\times$ the throughput of
eviction-only best-of-$N$, while full-KV runs out of memory. These results show that complementary
context views can preserve reasoning quality under severe KV memory
constraints.

% ===== 선택 섹션 (있으면 References 앞, 이 순서대로) =====================
% \section{Ethical Statement}   % 번호 없는 섹션
% \section*{Acknowledgments}
% =======================================================================
% ===== References ======================================================
% .bib 파일 이름을 적으세요 (확장자 제외). bibliographystyle 은 지정하지 마세요.
\bibliography{aaai2027}
% ===== Content Appendix (선택) =========================================

\appendix

\section{KV-Rescue vs KV Evicted Best-of-N}

\begin{figure*}[!t]
  \centering
  \includegraphics[width=0.8\textwidth]{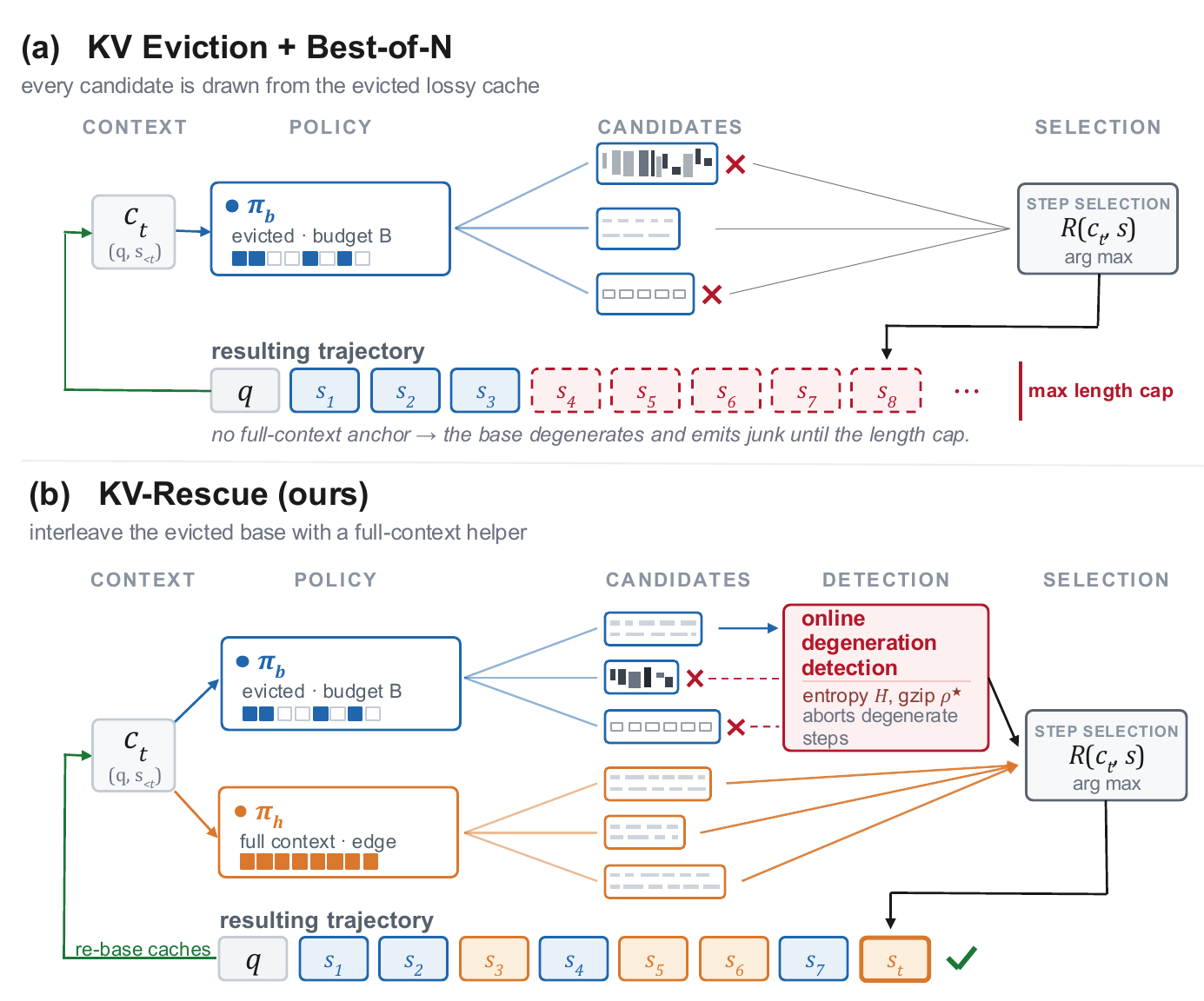}
  \caption{\textbf{KV-Rescue.} \textbf{(a)} Best-of-$N$ under KV eviction draws
  every candidate from the same lossy cache, so the information lost to eviction
  leads the base to produce wrong or degenerate steps. \textbf{(b)} KV-Rescue
  samples steps from an evicted base $\pi_b$ and a full-context helper $\pi_h$;
  an online degeneration detector prunes degenerate base steps, the PRM commits
  the best step $s_t$, and base and helper steps interleave.}
  \label{fig:method-full}
\end{figure*}

Figure~\ref{fig:method-full} contrasts the two settings. Under KV eviction,
best-of-$N$ draws every candidate step from the same lossy cache, so all
candidates share the base model's missing context: the PRM can only select
among them, and sampling cannot recover the evicted information. As a result
the base answers incorrectly or degenerates, emitting incoherent or repetitive
steps until the length cap (Fig.~\ref{fig:method-full}a).

KV-Rescue instead samples candidate steps from two policies at each step, the
evicted base $\pi_b$ and a full-context helper $\pi_h$. An online degeneration
detector prunes degenerate base candidates before selection, the PRM commits
the best surviving step $s_t$, and both caches re-base onto $c_{t+1}$. Committed
base and helper steps interleave over the solution, each covering the other's
failure and recovering the correct trajectory (Fig.~\ref{fig:method-full}b).

\section{KV-Rescue Pseudocode}

Algorithm~\ref{alg:kvrescue} summarizes one decoding pass. Each step samples
$N_b$ candidates from the evicted base $\pi_b$ (on the compressed cache
$\mathcal{E}_B(c)$) and $N_h$ from the full-context helper $\pi_h$, aborting any
that fire the online detector $\mathrm{deg}(\cdot)$ mid-generation. The PRM
commits the best non-degenerate step $s_t$, which extends the context $c$; both caches then re-base onto the
shared prefix. The loop repeats until an \texttt{<eos>} step or the length cap
$L$.

\begin{algorithm}[t]
\caption{KV-Rescue decoding for one question}
\label{alg:kvrescue}
\begin{algorithmic}[1]
\REQUIRE question $q$; evicted base policy $\pi_b$ with budget $B$;
         full-context helper policy $\pi_h$; PRM $R$;
         candidate counts $N_b,N_h$ with $N_h \ge 1$;
         detector thresholds $\tau_{\mathrm{ent}},\tau_{\mathrm{cmp}}$;
         maximum output-token length $L$
\ENSURE solution $\mathbf{s}$

\STATE $c \gets q$;\quad $\mathbf{s}\gets ()$;\quad $t\gets 1$
\STATE initialize the compressed base cache $K_b \gets \mathcal{E}_B(q)$
       and full-context helper cache $K_h \gets q$

\WHILE{$\operatorname{len}_{\mathrm{tok}}(\mathbf{s}) < L$}

  \STATE generate base candidates
  $\{s_t^{b,i}\}_{i=1}^{N_b}\sim\pi_b(\cdot\mid K_b)$,
  aborting a candidate as soon as it satisfies
  $\mathrm{deg}(s_t^{b,i};
  \tau_{\mathrm{ent}},\tau_{\mathrm{cmp}})$

  \STATE generate helper candidates
  $\{s_t^{h,j}\}_{j=1}^{N_h}\sim\pi_h(\cdot\mid K_h)$
  without degeneration screening

  \STATE $\widehat{\mathcal{C}}_t^b
  \gets
  \{\,s_t^{b,i} :
  \neg\mathrm{deg}(s_t^{b,i};
  \tau_{\mathrm{ent}},\tau_{\mathrm{cmp}})\,\}$

  \STATE $\mathcal{C}_t
  \gets
  \widehat{\mathcal{C}}_t^b
  \cup
  \{s_t^{h,j}\}_{j=1}^{N_h}$

  \STATE $s_t
  \gets
  \operatorname*{arg\,max}_{s\in\mathcal{C}_t} R(c,s)$
  \COMMENT{PRM selects the next step}

  \STATE $\mathbf{s}\gets\mathbf{s}\Vert s_t$;\quad
         $c\gets(c,s_t)$

  \IF{$s_t\in\widehat{\mathcal{C}}_t^b$}
    \STATE retain the selected base fork as $K_b$
    \STATE append $s_t$ to $K_h$
  \ELSE
    \STATE retain the selected helper fork as $K_h$
    \STATE append $s_t$ to $K_b$ and apply $\mathcal{E}_B$
  \ENDIF

  \IF{$s_t$ ends with \texttt{<eos>}}
    \STATE \textbf{break}
  \ENDIF

  \STATE $t\gets t+1$
\ENDWHILE

\STATE \textbf{return} $\mathbf{s}$
\end{algorithmic}
\end{algorithm}

\section{Implementation Details}
\label{sec:impl}

\paragraph{Models.}
We instantiate KV-Rescue on the Qwen2.5-Math family~\citep{qwen25math}. The \emph{base} reasoner is
\texttt{Qwen2.5-Math-7B-Instruct} (and \texttt{Qwen2.5-Math-72B-Instruct} for the scaling study); the full-KV
\emph{helper} is \texttt{Qwen2.5-Math-1.5B-Instruct}. Step selection uses the process reward model
\texttt{Qwen2.5-Math-PRM-7B}~\citep{zhang2025lessons}, a token-classification reward model: we form
$\text{prompt}\oplus\texttt{\textbackslash n\textbackslash n}\oplus s_1\oplus\dots\oplus s_k\oplus\texttt{\textbackslash n\textbackslash n}$
and take the softmax probability of the positive label at the final token as the step score $R(\cdot)$ (the
step separator is \texttt{\textbackslash n\textbackslash n}, matching the PRM's training-time separator). All
models run in \texttt{bfloat16}.

\paragraph{KV eviction.}
Eviction is applied to the base model only, via R-KV~\citep{cai2025rkv}, which compresses the base's KV to
a fixed budget $B$ at every step by scoring cached tokens with a mix of query-attention importance and key redundancy
(cosine similarity), keeping the top-$B$ plus a small recent window. We use budgets
$B\in\{64,128,256,512\}$, a headroom buffer $w{=}128$ (so the physical KV is bounded by $B'{=}B{+}w$), and mix
weight $\lambda{=}0.1$. The helper and PRM keep their full KV cache.

\paragraph{Best-of-$(N{+}N)$ trajectory.}
KV-Rescue grows a single accepted trajectory with a fork-based engine on top of vLLM~v1~\citep{kwon2023efficient} that
avoids re-prefilling history: one persistent sequence holds the accepted trajectory's (evicted) base KV, and
each step (i)~forks $N$ base candidates by physically copying that sequence's live (post-eviction) KV slots into
fresh blocks, (ii)~requests $N$ full-KV helper candidates in parallel, (iii)~scores all $2N$ candidates with the
PRM and accepts the best. If a base candidate is accepted it \emph{becomes} the trajectory's
new state (its KV already contains the step, evicted correctly); if a helper step is accepted, the base model
encodes its tokens in a single forward pass to extend the trajectory, so the base KV absorbs the step and
eviction stays incremental. The whole
procedure is forward-passes only, with no backtracking and no history re-prefill. We use $N{=}8$.
The engine is \emph{continuous-batched} and event-driven: rather than lock-stepping one problem at a time, it
keeps many problems in flight in a single shared running set, and a single \texttt{step()} advances all their
live candidates together (vLLM continuous-batches at the token level). Each problem proceeds to scoring the instant
its own $N$ base candidates have reached the step separator, and retires and is refilled independently, so fast
problems do not wait on the slowest and the base GPU stays saturated across problems. The per-step PRM and
helper calls (below) run on background threads, off the engine's critical path.

\paragraph{Online degeneration detection.}
Each base candidate is screened before scoring: it is dropped if its mean per-token
predictive entropy exceeds $\tau_{\mathrm{ent}}{=}7.0$ (incoherent generation) or the minimum gzip ratio of a
sliding window of width $w_{\mathrm{cmp}}{=}800$ falls below $\tau_{\mathrm{cmp}}{=}0.18$ (repetition loop); the
entropy check uses a guard of $24$ tokens and a window of $12$. Both signals are by-products of the existing
decode and add negligible cost.

\paragraph{Decoding and evaluation.}
Following the Qwen2.5-Math non-thinking regime we sample at temperature $0.8$, top-$p$ $0.95$, top-$k$ $20$,
with maximum context $4096$, up to $1024$ reasoning steps, and at most $512$ tokens per step. We evaluate
pass@1 accuracy on MATH500, AIME24, AMC23, GSM8K, and OlympiadBench, extracting the final \texttt{\textbackslash boxed} answer with
the official Qwen2.5-Math grader; AIME24 and AMC23 are averaged over five seeds and the rest use a single seed.

\paragraph{Serving.}
The base engine, PRM, and helper run as separate vLLM instances; the PRM and helper are queried over
asynchronous HTTP and overlap with base decoding, so their latency is hidden behind the base forward. For the
7B base we co-locate all three on each GPU and shard the problem set across GPUs with work-stealing; for the
72B base we shard the base model with tensor parallelism ($\text{TP}{=}8$) and co-locate the PRM
($\text{TP}{=}4$) and helper alongside it.

\section{Overhead Derivation}
\label{app:overhead}

We derive the per-sequence KV memory and per-token decode compute shown in
Fig.8, and the context length at which evicted decode becomes cheaper than
full-KV, and evaluate them with the Qwen2.5-Math constants.

\subsection{Notation}
For a decoder-only transformer let $L$ be the number of layers, $d$ the model dimension, $h_{\mathrm{kv}}$ the
number of key/value heads, $d_h$ the head dimension, $P$ the parameter count, and $b$ the bytes per element
($b{=}2$ for \texttt{bfloat16}). Let $\ell$ be the current context length (tokens), $B$ the R-KV budget,
$w$ the eviction buffer, and $B'{=}B{+}w$ the resulting physical KV bound. Table~\ref{tab:app-consts} lists the
values we use.

\begin{table}[t]
\centering\small
\caption{Model constants used in the derivation (from each model's \texttt{config.json}).}
\label{tab:app-consts}
\resizebox{\columnwidth}{!}{%
\begin{tabular}{lcccccc}
\toprule
model & $L$ & $d$ & $h_{\mathrm{kv}}$ & $d_h$ & $d_{\mathrm{ffn}}$ & $P$ \\
\midrule
base 7B  & 28 & 3584 & 4 & 128 & 18944 & $7.62\times10^{9}$ \\
base 72B & 80 & 8192 & 8 & 128 & 29568 & $72.7\times10^{9}$ \\
helper 1.5B & 28 & 1536 & 2 & 128 & 8960 & $1.54\times10^{9}$ \\
\bottomrule
\end{tabular}}
\end{table}

\subsection{KV memory}
Each cached token stores a key and a value vector of size $d_h$ in every KV head of every layer, so the
per-token, per-sequence KV footprint is
\begin{equation}
  m_{\mathrm{kv}} \;=\; 2\,L\,h_{\mathrm{kv}}\,d_h\,b,
  \label{eq:kvtok}
\end{equation}
the leading $2$ counting keys and values. Substituting Table~\ref{tab:app-consts} gives
$m_{\mathrm{kv}}^{\text{7B}}=56\,\text{KiB}$, $m_{\mathrm{kv}}^{\text{72B}}=320\,\text{KiB}$, and
$m_{\mathrm{kv}}^{\text{1.5B}}=28\,\text{KiB}$ per token.

Full-KV memory grows linearly in context, whereas eviction bounds it at the budget:
\begin{equation}
  M_{\mathrm{full}}(\ell)=m_{\mathrm{kv}}\,\ell,
  \qquad
  M_{\mathrm{evict}}=m_{\mathrm{kv}}\,B' \quad(\text{constant in }\ell).
  \label{eq:kvseq}
\end{equation}
The eviction saving on the base model is therefore \emph{model-size independent},
\begin{equation}
  \frac{M_{\mathrm{full}}(\ell)}{M_{\mathrm{evict}}}=\frac{\ell}{B'},
  \label{eq:kvratio}
\end{equation}
e.g.\ $\ell/B'{=}8192/640\approx12.8\times$ at $\ell{=}8\text{k}$, $B{=}512$. For the full method a problem
generates $N$ candidate sequences; the base is evicted and the helper keeps full KV, so
\begin{equation}
\begin{aligned}
  M_{\mathrm{KVR}} &= N\big(m_{\mathrm{kv}}^{\mathrm{base}}B' + m_{\mathrm{kv}}^{\mathrm{help}}\,\ell\big),\\
  M_{\mathrm{full\text{-}BoN}} &= N\,m_{\mathrm{kv}}^{\mathrm{base}}\,\ell .
\end{aligned}
\label{eq:kvproblem}
\end{equation}
The helper term $m_{\mathrm{kv}}^{\mathrm{help}}\ell$ does not depend on the base model, so it dominates
$M_{\mathrm{KVR}}$ when the base is small and is negligible when the base is large. At $\ell{=}4\text{k}$,
$B{=}512$, $N{=}8$, the ratio $M_{\mathrm{full\text{-}BoN}}/M_{\mathrm{KVR}}$ is $1.75/1.15\approx1.5\times$ for
the 7B base but $10.0/2.44\approx4.1\times$ for the 72B base.

\subsection{Decode compute}
For one decoded token the forward cost splits into a context-independent dense part (all linear projections and
the FFN, $\approx 2P$ FLOP by the standard multiply--add count) and an attention part that is linear in the
number of cached tokens attended to:
\begin{equation}
  C(\ell)=\underbrace{2P}_{C_{\mathrm{dense}}}\;+\;\underbrace{4\,L\,d\,\ell}_{C_{\mathrm{attn}}(\ell)} .
  \label{eq:compute}
\end{equation}
The attention term is the two matmuls per layer, $QK^\top$ and the value aggregation; each runs over the $\ell$
cached positions across the full query width $d$, costing $d\ell$ multiply--adds per token, so at two FLOP per
multiply--add the two matmuls give $4d\ell$ FLOP, summed over $L$ layers. (Under grouped-query attention the KV
\emph{memory} of Eq.~\eqref{eq:kvtok} scales with the $h_{\mathrm{kv}}$ cached KV heads, whereas attention
\emph{compute} scales with the full query width $d$.) With Table~\ref{tab:app-consts}, $C_{\mathrm{dense}}$ is
$15.2$ and $145.4$ GFLOP/token for 7B and 72B,
while $C_{\mathrm{attn}}(\ell)=0.401\text{M}\cdot\ell$ and $2.62\text{M}\cdot\ell$ FLOP respectively; at
$\ell{=}4\text{k}$ attention is only $11\%$ (7B) / $7\%$ (72B) of the forward, so bounded-KV attention saves
little \emph{compute} at short context.

Eviction is not free, however. When R-KV fires (every step once $\ell>B'$), it makes, per layer, one
attention-score pass plus a key-similarity, pooling and top-$k$ selection over the $B'$ physical tokens; to
first order this is $\alpha\,C_{\mathrm{attn}}(B')$ with $\alpha\in[1.5,3]$, so the evicted per-token cost
\begin{equation}
  C_{\mathrm{evict}}=C_{\mathrm{dense}}+(1{+}\alpha)\,C_{\mathrm{attn}}(B')=2P+(1{+}\alpha)\,4LdB'
  \label{eq:compute-evict}
\end{equation}
depends only on the bounded $B'$ and is thus flat in $\ell$. Evicted decode is cheaper per token once the
full-context attention outgrows this bounded cost, $C_{\mathrm{attn}}(\ell)\ge(1{+}\alpha)\,C_{\mathrm{attn}}(B')$.
As $C_{\mathrm{attn}}$ is linear in $\ell$, this happens beyond a crossover that is independent of model size,
\begin{equation}
  \ell^\star=(1{+}\alpha)\,B'.
  \label{eq:crossover}
\end{equation}
We measure this crossover at $\ell^\star\approx2500$; for $B'{=}640$ that corresponds to $\alpha\approx2.9$,
near the top of the $[1.5,3]$ range and consistent with R-KV doing more than a single attention pass (it also
computes the key similarity, pooling and top-$k$). Below $\ell^\star$ eviction is slightly slower, as the
scoring outweighs the attention it saves; above it eviction is faster, and the gap grows with $\ell-\ell^\star$,
reaching about $2\%$ at $4\text{k}$ and $6\%$ at $12\text{k}$. The $N$ helper candidates add, per token, a fraction
$\rho=C^{\mathrm{help}}(\ell)/C^{\mathrm{base}}_{\mathrm{evict}}$ of a base candidate,
\begin{equation}
  \rho=\frac{2P^{\mathrm{help}}+4L^{\mathrm{help}}d^{\mathrm{help}}\ell}
            {2P^{\mathrm{base}}+(1{+}\alpha)\,4L^{\mathrm{base}}d^{\mathrm{base}}B'} ;
  \label{eq:helper}
\end{equation}
the helper is a fixed $1.5$B model, so this fraction falls as the base grows, from about $24\%$ for the 7B
base to $2.5\%$ for the 72B base at $\ell{=}4\text{k}$.

These are the quantities in Fig.8: eviction caps the base KV cache (an $\ell/B'$
saving, Eq.~\eqref{eq:kvratio}) and its attention cost, at the price of a per-token overhead that is only
repaid past $\ell^\star\approx2500$ (Eq.~\eqref{eq:crossover}), plus the small helper cost $\rho$
(Eq.~\eqref{eq:helper}) of roughly $24\%$ for the 7B base and $2.5\%$ for the 72B base.

\begin{figure*}[t]
    \centering
    \includegraphics[width=\linewidth]{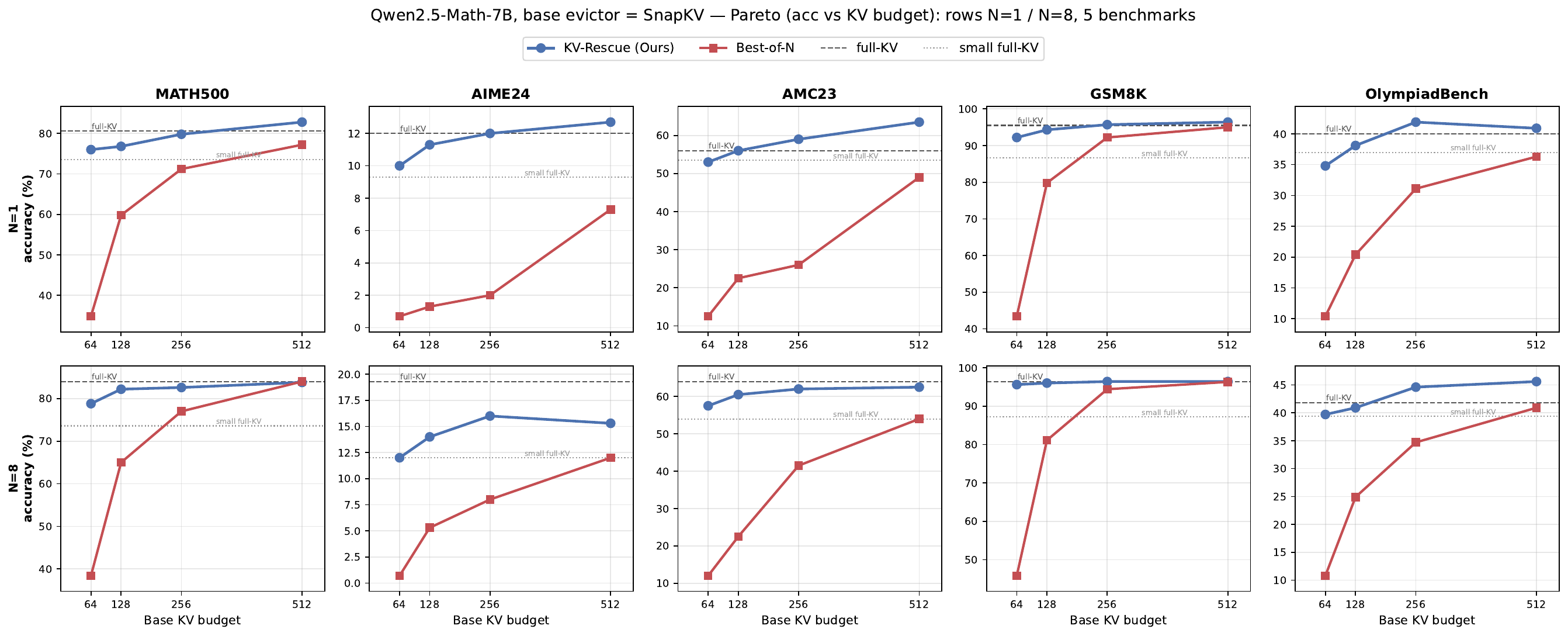}
    \caption{\textbf{KV-Rescue with SnapKV eviction (Qwen2.5-Math-7B).} pass@1 vs.\
base KV budget over the five benchmarks (MATH500, AIME24, AMC23, GSM8K,
OlympiadBench), for $N{=}1$ (top) and $N{=}8$ (bottom). Blue is KV-Rescue (ours),
red is eviction-only best-of-$N$; dashed and dotted lines mark the base full-KV
and helper (small) full-KV references. Same setup as main results
with R-KV replaced by SnapKV.}
    \label{fig:snapkv_pareto}
\end{figure*}

\begin{table*}[t]
\centering
\caption{Accuracy (\%, boxed+correct; mean$\pm$std over seeds) vs.\ base KV budget, for the $4\times5$ Pareto grid (Qwen2.5-Math 7B/72B, $N{=}1,8$). bon = evicted-main best-of-$N$; \textbf{b2n} = ours (+full-KV 1.5B safety net); full-KV / small full-KV are no-eviction references. Metric identical to the Pareto figure (is\_correct $\wedge$ \texttt{\textbackslash boxed}; PRM/small-degraded rows excluded). std shown only where $>$1 seed (AIME, AMC23).}
\label{tab:main_results}
\scriptsize\centering{
\begin{tabular}{@{}lllccccc@{}}
\toprule
model & $N$ & budget & MATH500 & AIME24 & AMC23 & GSM8K & Olympiad \\
\midrule
\multirow{10}{*}{7B} & \multirow{10}{*}{1} & bon 64 & 46.8 & 0.7\,\tiny{$\pm$1.3} & 16.5\,\tiny{$\pm$3.7} & 54.4 & 13.8 \\
 & & bon 128 & 64.4 & 1.3\,\tiny{$\pm$1.6} & 28.0\,\tiny{$\pm$2.9} & 81.0 & 25.8 \\
 & & bon 256 & 73.6 & 4.7\,\tiny{$\pm$1.6} & 46.0\,\tiny{$\pm$9.7} & 92.0 & 32.9 \\
 & & bon 512 & 77.0 & 4.7\,\tiny{$\pm$2.7} & 50.0\,\tiny{$\pm$6.5} & 95.3 & 36.1 \\
\cmidrule(l){3-8}
 & & \textbf{b2n} 64 & 77.8 & 12.7\,\tiny{$\pm$3.9} & 58.0\,\tiny{$\pm$5.6} & 91.8 & 38.4 \\
 & & \textbf{b2n} 128 & 78.6 & 12.7\,\tiny{$\pm$1.3} & 60.0\,\tiny{$\pm$3.5} & 94.2 & 39.4 \\
 & & \textbf{b2n} 256 & 80.8 & 14.7\,\tiny{$\pm$2.7} & 62.5\,\tiny{$\pm$1.6} & 95.0 & 41.2 \\
 & & \textbf{b2n} 512 & 83.0 & 16.7\,\tiny{$\pm$4.7} & 64.0\,\tiny{$\pm$2.5} & 95.9 & 42.5 \\
 & & full-KV & 80.6 & 12.0\,\tiny{$\pm$5.4} & 56.0\,\tiny{$\pm$3.0} & 95.5 & 40.0 \\
 & & small full-KV & 73.6 & 9.3\,\tiny{$\pm$2.5} & 53.5\,\tiny{$\pm$4.1} & 86.6 & 37.0 \\
\midrule
\multirow{10}{*}{7B} & \multirow{10}{*}{8} & bon 64 & 52.8 & 1.3\,\tiny{$\pm$1.6} & 19.0\,\tiny{$\pm$3.0} & 61.0 & 16.7 \\
 & & bon 128 & 69.2 & 4.7\,\tiny{$\pm$1.6} & 44.5\,\tiny{$\pm$5.6} & 82.9 & 29.6 \\
 & & bon 256 & 79.4 & 8.7\,\tiny{$\pm$2.7} & 54.0\,\tiny{$\pm$8.5} & 94.7 & 38.1 \\
 & & bon 512 & 85.2 & 14.0\,\tiny{$\pm$4.9} & 60.0\,\tiny{$\pm$4.5} & 96.0 & 41.6 \\
\cmidrule(l){3-8}
 & & \textbf{b2n} 64 & 82.2 & 12.0\,\tiny{$\pm$2.7} & 60.5\,\tiny{$\pm$6.2} & 94.8 & 40.6 \\
 & & \textbf{b2n} 128 & 83.2 & 14.0\,\tiny{$\pm$4.9} & 62.0\,\tiny{$\pm$3.7} & 95.9 & 40.9 \\
 & & \textbf{b2n} 256 & 84.2 & 14.7\,\tiny{$\pm$1.6} & 63.0\,\tiny{$\pm$4.0} & 96.6 & 42.8 \\
 & & \textbf{b2n} 512 & 85.4 & 16.0\,\tiny{$\pm$2.5} & 62.0\,\tiny{$\pm$5.8} & 96.8 & 44.0 \\
 & & full-KV & 84.0 & 19.3\,\tiny{$\pm$2.5} & 64.0\,\tiny{$\pm$2.5} & 96.4 & 41.8 \\
 & & small full-KV & 73.6 & 12.0\,\tiny{$\pm$2.7} & 54.0\,\tiny{$\pm$3.7} & 87.2 & 39.4 \\
\midrule
\multirow{10}{*}{72B} & \multirow{10}{*}{1} & bon 64 & 36.0 & 2.0\,\tiny{$\pm$2.7} & 10.5\,\tiny{$\pm$1.9} & 42.5 & 10.2 \\
 & & bon 128 & 53.8 & 4.0\,\tiny{$\pm$3.3} & 23.5\,\tiny{$\pm$3.7} & 67.2 & 19.1 \\
 & & bon 256 & 68.6 & 5.3\,\tiny{$\pm$2.7} & 38.5\,\tiny{$\pm$5.1} & 88.5 & 27.6 \\
 & & bon 512 & 78.6 & 12.0\,\tiny{$\pm$4.0} & 48.5\,\tiny{$\pm$6.4} & 93.0 & 38.8 \\
\cmidrule(l){3-8}
 & & \textbf{b2n} 64 & 76.4 & 12.0\,\tiny{$\pm$1.6} & 55.0\,\tiny{$\pm$7.7} & 91.3 & 39.3 \\
 & & \textbf{b2n} 128 & 79.2 & 12.0\,\tiny{$\pm$4.5} & 58.5\,\tiny{$\pm$3.7} & 93.6 & 39.7 \\
 & & \textbf{b2n} 256 & 80.6 & 14.7\,\tiny{$\pm$5.0} & 63.5\,\tiny{$\pm$6.6} & 95.1 & 41.0 \\
 & & \textbf{b2n} 512 & 83.4 & 16.0\,\tiny{$\pm$4.4} & 63.5\,\tiny{$\pm$3.4} & 96.1 & 43.0 \\
 & & full-KV & 82.6 & 15.3\,\tiny{$\pm$2.7} & 56.5\,\tiny{$\pm$4.1} & 93.3 & 43.3 \\
 & & small full-KV & 73.6 & 9.3\,\tiny{$\pm$2.5} & 53.5\,\tiny{$\pm$4.1} & 86.6 & 37.0 \\
\midrule
\multirow{10}{*}{72B} & \multirow{10}{*}{8} & bon 64 & 60.6 & 4.0\,\tiny{$\pm$1.3} & 31.5\,\tiny{$\pm$6.0} & 74.1 & 28.6 \\
 & & bon 128 & 74.8 & 9.3\,\tiny{$\pm$3.3} & 52.0\,\tiny{$\pm$5.3} & 90.5 & 35.7 \\
 & & bon 256 & 83.4 & 14.0\,\tiny{$\pm$1.3} & 62.0\,\tiny{$\pm$3.3} & 95.0 & 43.3 \\
 & & bon 512 & 85.6 & 19.3\,\tiny{$\pm$6.5} & 69.5\,\tiny{$\pm$5.1} & 96.2 & 46.8 \\
\cmidrule(l){3-8}
 & & \textbf{b2n} 64 & 82.6 & 15.3\,\tiny{$\pm$3.4} & 61.0\,\tiny{$\pm$5.1} & 95.2 & 43.4 \\
 & & \textbf{b2n} 128 & 82.4 & 18.0\,\tiny{$\pm$2.7} & 61.5\,\tiny{$\pm$4.6} & 96.0 & 43.7 \\
 & & \textbf{b2n} 256 & 86.2 & 17.3\,\tiny{$\pm$2.5} & 67.0\,\tiny{$\pm$4.3} & 96.3 & 46.7 \\
 & & \textbf{b2n} 512 & 86.6 & 19.3\,\tiny{$\pm$3.3} & 72.5\,\tiny{$\pm$8.5} & 97.0 & 48.4 \\
 & & full-KV & 86.4 & 23.3\,\tiny{$\pm$3.7} & 71.5\,\tiny{$\pm$3.4} & 96.5 & 50.2 \\
 & & small full-KV & 73.6 & 12.0\,\tiny{$\pm$2.7} & 54.0\,\tiny{$\pm$3.7} & 87.2 & 39.4 \\
\bottomrule
\end{tabular}}
\end{table*}

\section{Experiment Results}

\subsection{Main Result}

Table~\ref{tab:main_results} provides the exact pass@1 numbers behind the main
results figure across the $4\times5$ grid.

\subsection{KV-Rescue with SnapKV}

KV-Rescue is agnostic to the underlying evictor: it treats the base as a
black-box policy conditioned on whatever cache the evictor keeps. To verify this,
Figure~\ref{fig:snapkv_pareto} repeats the 7B study with R-KV replaced by
SnapKV~\citep{li2024snapkv}, and the same pattern holds, showing the recovery
does not depend on the specific eviction method.

\end{document}